\documentclass[10pt,twocolumn,letterpaper]{article}
\usepackage[table,xcdraw,dvipsnames]{xcolor}
\usepackage{makecell}
\usepackage[algo2e]{algorithm2e}
\usepackage{algorithm}
\usepackage{listings}
\usepackage{multirow}
\usepackage{soul}
\usepackage{wrapfig}
\usepackage{amsmath}
\usepackage{bm}
\usepackage{tabularx}

\usepackage[pagenumbers]{cvpr} 

\definecolor{cvprblue}{rgb}{0.21,0.49,0.74}
\usepackage[pagebackref,breaklinks,colorlinks,allcolors=cvprblue]{hyperref}

\def\paperID{***} 
\def\confName{CVPR}
\def\confYear{2025}

\title{Exploring CLIP's Dense Knowledge for Weakly Supervised Semantic Segmentation}

\author{
Zhiwei Yang$^{1,2}$\qquad Yucong Meng$^{2,3}$\qquad\\
Kexue Fu$^{4}$\qquad Feilong Tang$^{1}$\qquad Shuo Wang$^{2,3\thanks{Corresponding authors. Email:\{shuowang, zjsong\}@fudan.edu.cn.}}$\qquad Zhijian Song$^{1,2,3\footnotemark[1]}$ \\
$^{1}$Academy for Engineering and Technology, Fudan University, Shanghai 200433, China\\
$^{2}$Shanghai Key Laboratory of Medical Image Computing and Computer Assisted Intervention\\
$^{3}$Digital Medical Research Center, School of Basic Medical Sciences, Fudan University, China\\
$^{4}$Shandong Computer Science Center (National Supercomputer Center in Jinan)
}

\newcommand{\tablestyle}[2]{\setlength{\tabcolsep}{#1}\renewcommand{\arraystretch}{#2}\centering\footnotesize}

\begin{document}
\maketitle
\begin{abstract}
    Weakly Supervised Semantic Segmentation (WSSS) with image-level labels aims to achieve pixel-level predictions using Class Activation Maps (CAMs). Recently, Contrastive Language-Image Pre-training (CLIP) has been introduced in WSSS. However, recent methods primarily focus on image-text alignment for CAM generation, while CLIP's potential in patch-text alignment remains unexplored. In this work, we propose ExCEL to explore CLIP's dense knowledge via a novel patch-text alignment paradigm for WSSS. Specifically, we propose Text Semantic Enrichment (TSE) and Visual Calibration (VC) modules to improve the dense alignment across both text and vision modalities. To make text embeddings semantically informative, our TSE module applies Large Language Models (LLMs) to build a dataset-wide knowledge base and enriches the text representations with an implicit attribute-hunting process. To mine fine-grained knowledge from visual features, our VC module first proposes Static Visual Calibration (SVC) to propagate fine-grained knowledge in a non-parametric manner. Then Learnable Visual Calibration (LVC) is further proposed to dynamically shift the frozen features towards distributions with diverse semantics. With these enhancements, ExCEL not only retains CLIP's training-free advantages but also significantly outperforms other state-of-the-art methods with much less training cost on PASCAL VOC and MS COCO. Code is available at \href{https://github.com/zwyang6/ExCEL}{https://github.com/zwyang6/ExCEL}.
\end{abstract}

\vspace{-1.em}
\section{Introduction}
\label{sec:intro}

Weakly Supervised Semantic Segmentation (WSSS) intends to generate pixel-level predictions using weak annotations like points~\cite{1}, scribbles~\cite{2,3}, bounding boxes~\cite{4,5}, or image-level labels~\cite{6,7,8}. It significantly reduces the annotating cost of fully supervised methods and has attracted increasing attention in the community. Among all cheap annotation types, most WSSS approaches leverage image-level labels to provide dense localization cues, linking visual concepts to specific pixel regions~\cite{9,32}. In this work, we focus on WSSS with image-level labels as well.

\begin{figure}[t!]
  \centering
  \includegraphics[width=8.36cm]{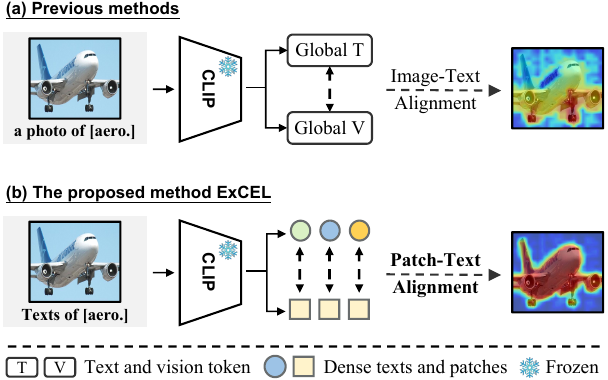}
   \caption{Our motivation. (a) Previous methods leverage CLIP to generate CAMs with global image-text alignment, leaving CLIP's dense knowledge unexplored. (b) The proposed ExCEL explores CLIP's dense knowledge via a novel patch-text alignment paradigm, which generates better CAMs with less training cost.}
   \label{fig.1}
   \vspace{-1.2em}
\end{figure}

Commonly, the WSSS pipeline involves three stages: generating Class Activation Maps (CAMs)~\cite{11} by training a classification network, refining CAMs into pseudo labels~\cite{12}, and using these labels to train a segmentation model~\cite{OCR}. However, due to the minimal semantic information from image-level labels, CAMs intend to highlight the most distinctive object parts, significantly limiting WSSS performance. Recently, Contrastive Language-Image Pre-training (CLIP)~\cite{14} has been introduced in WSSS. CLIMs~\cite{15} applies image-text pairs to regularize visual relations among different semantics. CLIP-ES~\cite{16} leverages image-text alignment for gradient and produces high-quality GradCAM~\cite{17}. WeCLIP~\cite{18} further streamlines this process by using CLIP's visual encoder for segmentation. Despite these advancements, current methods primarily focus on CLIP's global image-text alignment, as shown in \cref{fig.1} (a). CLIP's dense knowledge with patch-text alignment still remains under-explored in WSSS.

In this work, we propose ExCEL to explore CLIP's dense knowledge via a patch-text alignment paradigm for WSSS, i.e., generating CAMs by calculating patch-wise similarity between text and individual patch tokens, as shown in \cref{fig.1} (b). We identify two key challenges: (1) Semantic sparsity in textual prompts, where the template 'a photo of [CLASS]' only indicates object presence but lacks knowledge for localization, and (2) Fine-grained insufficiency in visual features, as CLIP prioritizes global representation due to its image-text pairing nature. To address these, ExCEL enhances CLIP's dense alignment with Text Semantic Enrichment (TSE) and Visual Calibration (VC) modules, unlocking its potential across text and vision modalities.

To generate semantically rich text representations, we propose TSE through an implicit attribute feature space. Instead of relying on explicit text templates, TSE module implicitly constructs text embeddings using universal attributes across the dataset. We first employ Large Language Models (LLMs) to generate detailed descriptions for each class, which are then processed by CLIP's text encoder to build a dataset-wide knowledge base. Rather than directly fusing these class-specific descriptions for text prompting, we focus on clustering the descriptive embeddings into generalized attributes. which effectively capture complementary knowledge from other classes and supplement missing information for the target class. With this implicit feature space, we enhance the text embeddings by hunting for its most relevant attribute features and aggregate them into the final class-specific text representation. This approach enables TSE to generate more informative text embeddings, providing a strong foundation for visual recognition.

To mine fine-grained knowledge from visual features, we propose VC to calibrate CLIP in both non-trainable and efficient-learnable ways. Our findings suggest that CLIP's q-k attention loses fine-grained details. Therefore, we first propose a Static Visual Calibration (SVC) module to replace the suboptimal q-k attention with a straightforward Intra-correlation operation. It focuses on extracting fine-grained details from intermediate layers, which progressively propagates fine-grained visual knowledge. Without any retraining, SVC generates CAMs comparable to training-required WSSS methods. Building on this, we further propose a Learnable Visual Calibration (LVC) module to dynamically calibrate CLIP's frozen features. LVC extracts spatial correlations from SVC’s static CAMs. These correlations further supervise a lightweight adapter to learn the dynamic shift, pushing frozen features towards spatial-aware distributions. LVC and SVC complement each other, enabling precise patch-text alignment for CAM generation.

The main contributions of our work are listed as follows:
\begin{itemize}
    \item We explore CLIP's dense knowledge via a novel patch-text alignment paradigm for WSSS. The proposed ExCEL generates better pseudo labels in both training-free and efficient learning manners, revealing the dense capabilities of CLIP for efficient CAM generation.
    \item To enhance patch-text alignment, we propose the Text Semantic Enrichment (TSE) and Visual Calibration (VC) modules. TSE applies LLMs to build a dataset-wide knowledge base and treats text prompting as an implicit attribute-hunting process, making text embeddings more informative. VC propagates the fine-grained visual knowledge in a non-parametric manner and further dynamically calibrates the frozen features with a lightweight adapter. TSE and VC work across two modalities, generating better dense alignment and pseudo labels.
    \item Extensive experiments on PASCAL VOC and MS COCO demonstrate that ExCEL significantly outperforms recent state-of-the-art methods, while reducing training cost with only $3.2$ GB of GPU memory and $6\%$ of the training time required by recent methods.
\end{itemize}

\section{Related Works}
\subsection{Weakly Supervised Semantic Segmentation}
\label{sec2.1}
Weakly supervised semantic segmentation with image-level labels typically relies on CAMs to provide dense supervision for segmentation~\cite{34,CGM}. However, CAMs usually highlight the most discriminative parts of objects~\cite{20}. To address this issue, considerable efforts have been made from many intriguing insights. MCTformer~\cite{21} incorporates multiple class tokens in Vision Transformer and proposes generating CAMs from class-patch attention. ToCo~\cite{22} proposes token contrast learning and generates more precise CAMs. SeCo~\cite{SeCo} designs a separate and conquer scheme and succeeds in tackling co-occurrence. Despite these advancements, prior methods commonly require retraining the entire classification network for CAM generation. In this work, our ExCEL directly generates CAMs from frozen CLIP, and further boosts its quality via a lightweight adapter, significantly reducing the training cost.
\begin{figure*}[t!]
  \centering    
  \includegraphics[width=17.2cm]{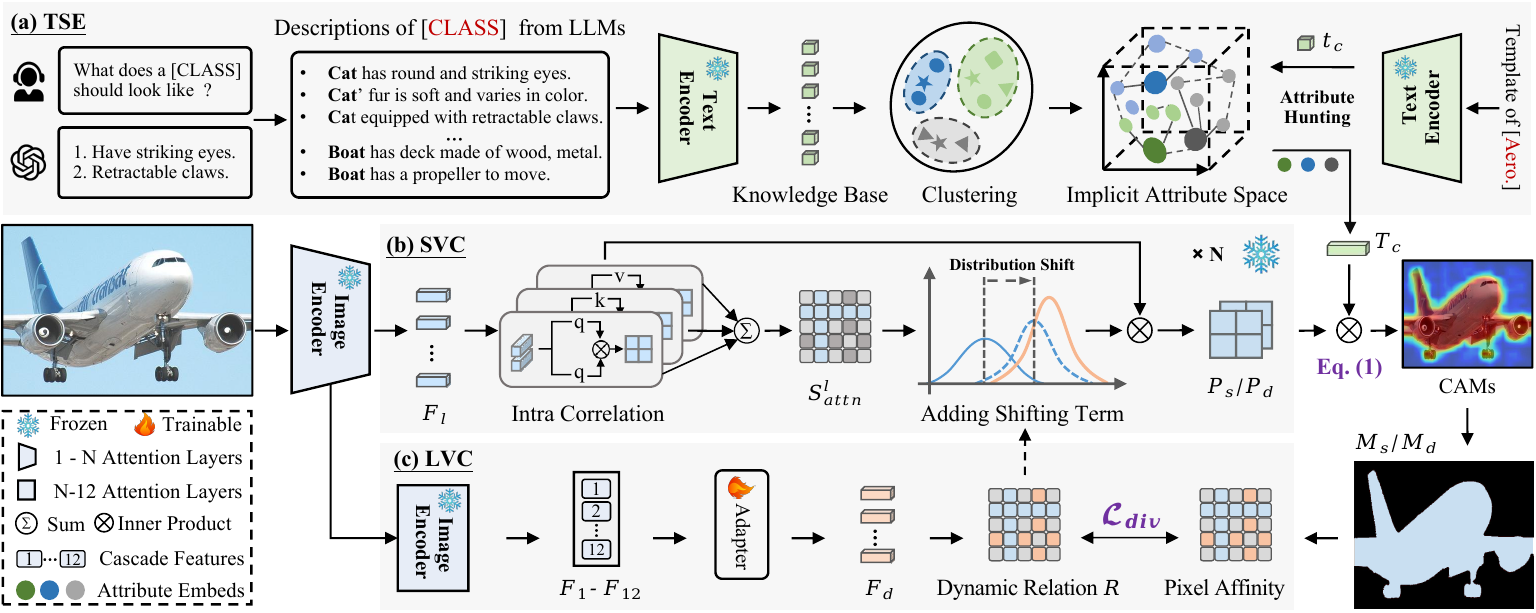}
   \caption{ExCEL Architecture. We explore CLIP's dense knowledge with Text Semantic Enrichment (TSE) and Visual Calibration (VC). (a) TSE uses LLMs to build a knowledge base and clusters it into an implicit attribute space. The final text representation $T_c$ is enhanced by hunting for relevant attributes. For vision modality, (b) we introduce Static Visual Calibration (SVC) to calibrate visual features using the Inter-correlation operation across $N$ intermediate layers. It generates static CAMs with $T_c$ and calibrated features $P_s$. (c) Learnable Visual Calibration (LVC) designs a learnable adapter to add a dynamic shift $R$ to SVC. It generates optimized features $P_d$ based on static CAMs guidance, creating dynamic CAMs from $P_d$ and $T_c$. Dynamic CAMs are refined for segmentation supervision. Details are in \cref{sec.3.1}.}
   \label{fig.2}
   \vspace{-1.em}
\end{figure*}

\subsection{Vision-Language Pre-training}
 Contrastive Language Image Pre-training (CLIP)~\cite{14}, known for pretraining on billion-scale image-text pairs, has demonstrated remarkable transferability in many downstream tasks. CoOP~\cite{24} and CLIP-Adapter~\cite{25} incorporate lightweight trainable parameters into CLIP and succeed in few-shot classification. DenseCLIP~\cite{26} and MaskCLIP~\cite{27} leverage the alignment between text and vision modalities for the dense segmentation task. Recently, some studies have introduced CLIP into WSSS. CLIMS~\cite{15} treats image-text pairing of CLIP as regularization and leverages it to regularize the visual concepts. CLIP-ES~\cite{16} finds that the image-text alignment of CLIP generates class gradient and leverages it for GradCAM generation. WeCLIP~\cite{18} further streamlines this process and directly leverages CLIP's visual encoder for segmentation. However, these methods mainly focus on a global image-text alignment while ignoring the dense capabilities of CLIP. In contrast, our ExCEL explores CLIP's dense knowledge via a patch-text alignment paradigm for WSSS.

\section{Methodology}
\subsection{Preliminaries}
\label{sec.3.1}
{\bf Patch-text CAM Generation.} CLIP uses image and text encoders to project image and text into the same feature space, enabling robust vision-language alignment. In this work, we utilize this property to generate CAM in a patch-text alignment paradigm. Given the encoded text embeddings $T \in \mathbb{R}^{D \times C}$ and visual features $P \in \mathbb{R}^{h \times w \times D}$, where $D$ is the feature channel, $h$ and $w$ are the spatial sizes, $C$ is the number of classes, we generate CAM by calculating the patch-wise similarities between text and visual features: 
\begin{equation}                            
\operatorname{CAM}=\operatorname{Norm}\left(\operatorname{cos}\left({P}, {T}\right),\right.
    \label{eq:0}
\end{equation}
where $\operatorname{Norm}(\cdot)$ is the min-max normalization and $\operatorname{cos}(\cdot)$ is the cosine similarity calculation. However, due to the image-level pairing nature, CLIP suffers from textual semantic sparsity and visual fine-grained insufficiency. Therefore, our ExCEL incorporates TSE and VC (SVC and LVC) into CLIP to further explore its dense potential. 

{\bf Framework Overview.} Our ExCEL generates CAMs in both training-free and efficient learning manners, as shown in \cref{fig.2}. Its training pipeline is generalized as follows: 

{(1) Enriching textual semantics via TSE.} We first use GPT-4 to generate descriptions for each class, which are encoded into a dataset-wide knowledge base with CLIP's text encoder. We cluster this knowledge into class-agnostic attributes and use the global text prompt to hunt for its most relevant ones. They are then aggregated into the final text representation. 
{(2) Static CAM generation via SVC.} We replace CLIP's q-k self-attention with our Intra-correlation operation from intermediate layers. Then the calibrated visual features and enhanced text embeddings are used for static CAMs via~\cref{eq:0}. 
{(3) Dynamic CAM generation via LVC.} A lightweight adapter is designed to learn dynamic token relations from static CAMs. The relations are added to SVC and serve as a distribution shift to make the visual features more diverse. The dynamic CAMs are generated with the enhanced text embeddings and LVC features via~\cref{eq:0}. 
{(4) Segmentation training.} Dynamic CAMs are refined to pseudo labels for segmentation supervision. 

\subsection{Text Semantic Enrichment}
\label{sec.3.2}
\textbf{Knowledge Base Construction.} The global text template ‘a clean origami of [CLASS]’ only indicates the presence of objects while limited providing dense knowledge for patch-wise visual recognition. To enrich the text representation, we first adapt LLMs, such as GPT-4 to generate detailed class descriptions, as shown in \cref{fig.2} (a). Specifically, given the global template $E_c$ from class label space $Y \in \{1,2,..., C\}$, where $c$ is the class index and $C$ is the number of categories, we carefully construct instructions for GPT: \textit{"List $n$ descriptions with key properties to describe the [CLASS] in terms of appearance, color, shape, size, or material, etc. These descriptions will help visually distinguish the [CLASS] from other classes in the dataset. Each description should follow the format: 'a clean origami [CLASS]. it + descriptive contexts.'"} With this instruction, we ask GPT to generate $n$ detailed descriptions for each class, which are subsequently encoded into a dataset-wide knowledge base with CLIP's text encoder. The knowledge base is denoted as $\mathcal{T}=\left\{\Phi\left(e_i\right)\right\}_{i=1}^{n \times C}$. $e_i$ is the description from GPT, $\Phi(\cdot)$ is CLIP's text encoder and $n\times C$ is the number of all descriptions. This knowledge base gathers descriptive properties for the whole dataset, building a strong foundation for the textual category representation.

\textbf{Implicit Attribute Hunting.} Instead of explicitly merging class-related knowledge into a single text embedding, we cluster this knowledge into generalized attributes and treat text prompting as an implicit attribute-hunting process. This has two main benefits: (1) $n$ explicit descriptions may still be limited in covering all characteristics of the class. The clustered attributes efficiently capture shared contextual knowledge from other categories, supplementing missing information for target class recognition. (2) The generated descriptions inevitably contain redundant or noisy content. The use of attributes makes the knowledge more compact and representative, leading to precise text prompting. 

To this end, we leverage a clustering algorithm to generate multiple centroids based on the knowledge base. Each cluster centroid is viewed as the implicit attribute that represents a group of descriptions sharing similar properties:
\begin{equation}                            
    {A}=\operatorname{Kmeans}(\mathcal{T}, B)=\left\{a_i\right\}_{i=1}^B,
    \label{eq:1}
\end{equation}
where $B$ is the number of centroids and kmeans algorithm~\cite{kmeans} is used for clustering for simplicity. $a_i \in \mathbb{R}^{D \times 1}$ represents the cluster centroid, i.e., the implicit attribute. 

With the attribute feature space $A$, we first send the global template $E_c$ into the text encoder of CLIP to generate global text embedding $t_c \in \mathbb{R}^{D \times 1}$. $D$ is the channel dimension. Then $t_c$ is leveraged to search for its most relevant attributes in the implicit attribute space. To exclude irrelevant attributes, we further propose to select TOPK attribute neighbors based on the similarity scores with $t_c$:
\begin{equation}                            
    A_c=\{a_j: j \in \operatorname{argmax}_{\text {TOPK }}\left\{t_c^T a_j\right\}_{j=1}^B\}.
    \label{eq:2}
\end{equation}

 Finally, we gently aggregate the implicit attribute neighbors according to the corresponding similarity weights to $t_c$ and take the aggregated features as the complementary knowledge for textual semantic enrichment. The final text representation $T_c \in \mathbb{R}^{D \times 1}$ is denoted as:
\begin{equation}                            
    {T}_{{c}}={t}_{{c}}+\lambda \sum_{{j}=1}^{{K}} \operatorname{softmax}\left({t}_{{c}}^{{T}} {~A}_{{c}}\right) {a}_{{j}},
    \label{eq:3}
\end{equation}
where $\lambda$ is the factor to balance the attribute information.

\subsection{Visual Calibrations}
\label{sec.3.4}

\textbf{Static Visual Calibration.} Due to the image-text pairing nature of CLIP, the visual features lack fine-grained information, leading to unreasonable localization maps via patch-text alignment. To delve into this, let us review the self-attention mechanism of the original CLIP first. As shown in~\cref{fig.2} (b), given the input image $X \in \mathbb{R}^{3\times \mathcal{H} \times \mathcal{W}}$, we send it to the image encoder of CLIP, which includes $12$ Attention layers in this work. $\mathcal{H} \times \mathcal{W}$ is the image size. For the feature $F_l \in \mathbb{R}^{D_s \times hw} $ from $l$-th layer of CLIP, it is first projected into three different spaces $\{q,k,v\}$, named query, key, and value, respectively. $q, k$ and $v$ have the same shape of ${D_s \times hw}$, where $D_s$ and $hw$ represents the channel dimension and sequence length. Then the attention map between $q$ and $k$ is calculated by measuring the similarity:
\begin{equation}                            
    \operatorname{SA}({q}, {k})=\operatorname{sofmax}\left({q}^{{T}} {k} / \sqrt{{D}_{{s}}}\right),
    \label{eq:4}
\end{equation}
where $\operatorname{SA}(\cdot)$ is the calculation of attention. Then the output features $F_{l+1}$ are generated by aggregating the tokens of $v$ according to similarity weights in the attention map. 

However, due to the inherent image-text alignment of CLIP, the original q-k attention produces overly uniform attention maps, homogenizing diverse tokens from $v$ to capture broad semantics for global image representation (see discussions in \cref{sec.4.4}). It leads to inaccurate object recognition. MaskCLIP~\cite{27} holds a similar observation, supporting this claim by removing the final q-k attention layer and using $v$ from the last layer as the visual output to preserve diversity. In our work, we choose to replace the suboptimal q-k attention with a straightforward Intra-correlation operation and focus on extracting fine-grained details from intermediate layers. This non-parametric approach effectively mines spatial semantics in intermediate $\{q,k,v\}$ and avoids the smoothing effect of q-k attention, resulting in more consistent attention maps and improved object localization.

Specifically, instead of generating q-k correlation, Intra-correlation calculates the attention within each space of $\{q,k,v\}$ across intermediate layers. The attention map ${S}_{{attn}}^l \in \mathbb{R}^{hw \times hw}$ from $l$-th SVC layer is generated by:
\begin{equation}                            
    {S}_{{attn}}^l=\sum {w}_{{i}} \operatorname{SA}\left({O}_{{i}}^{{l}}, {O}_{{i}}^{{l}}\right), {O}_{{i}}^{{l}} \in\left\{{q}^l, {k}^{{l}}, {v}^{{l}}\right\},
    \label{eq:5}
\end{equation}
where $w_i$ is the contribution weight for different correlation maps. $l \in \{12-N, ..., 12\}$ and $N$ is the number of intermediate layers for this operation. Then ${S_{attn}^l}$ and $v^l$ are used to generate the output features. Finally, the calibrated features $P_s \in \mathbb{R}^{D \times h \times w}$ from the last layer is used to generate static CAM ${CAM}_{{s}}$ with text embedding $T_c$ via~\cref{eq:0}.

\textbf{Learnable Visual Calibration.} Although ExCEL generates comparable CAMs without training, its performance is still limited by the fixed features in CLIP. To further unleash the dense potential of CLIP, we design a lightweight adapter to dynamically calibrate the visual features with diverse details. This adapter only incorporates a distribution shift to calibrate the fixed features without changing CLIP's pre-trained weights, thereby retaining CLIP's transferability and enhancing its dense performance for WSSS. 

Specifically, as shown in \cref{fig.2} (c), frozen features $F_l$ from $1$-$12$th layer of CLIP are extracted to learn a dynamic feature $F_d$ via the adapter. The process is expressed as:
\begin{equation}                            
{F}_{{d}}=\operatorname{Conv}(\operatorname{Concate}\left[\delta_l\left({F}_l\right)\right]_{l=1}^{12}),
    \label{eq:7}
\end{equation}
where $F_d \in \mathbb{R}^{D_d \times hw}$, $D_d$ is the channel dimension. $\operatorname{Conv}(\cdot)$ is the convolution layer, $\operatorname{Concate}[\cdot]$ is the concatenate operation that connects all features along channel dimension, and $\delta_l(\cdot)$ is the individual MLP layer for each $F_l$. Then $F_d$ is used to generate dynamic token relations by:
\begin{equation}                            
    r=\alpha(\operatorname{cos}\left(F_d, F_d\right)-\beta \overline{\operatorname{cos}\left(F_d, F_d\right)}),
    \label{eq:8}
\end{equation}
where $r \in \mathbb{R}^{hw \times hw}$, $\alpha$ and $\beta$ are the scaling and shifting factors to adjust the relations, respectively. $\overline{\operatorname{cos}(F_d, F_d)}$ means the mean value of similarity scores of $F_d$. It is designed to remove the irrelevant relations in low values by:
\begin{equation}                            
    R_{i j}= \begin{cases}r_{i j}, & \text { if } r_{i j} \geq 0 \\ -i n f, & \text { else }\end{cases}.
    \label{eq:9}
\end{equation}

With the dynamic relation $R \in \mathbb{R}^{hw \times hw}$, we add it as a distribution bias to the static attention map $S_{attn}^l$, dynamically grouping the frozen tokens within related semantics and shifting the features towards denser distribution. The optimized attention map $L_{attn}^l \in \mathbb{R}^{hw \times hw}$ is denoted as:
\begin{equation}                            
{L}_{{attn}}^{{l}}={S}_{{attn}}^{{l}}+\operatorname{softmax}({R}).
    \label{eq:10}
\end{equation}

 Subsequently, we extract the dynamically calibrated features $P_d \in \mathbb{R}^{D \times h \times w}$ from the last layer of LVC and generate dynamic CAMs with~\cref{eq:0}, which are then refined to final pseudo labels $M_d$ for the segmentation.

\subsection{Training Objectives}
\label{sec.3.3}
We formulate a diversity loss to supervise the learning of $F_d$ in our LVC module. We first measure token correlations of $F_d$ by calculating the self similarity: $\hat{\mathcal{R}} =\operatorname{sigmoid} (\operatorname{cos}(F_d, F_d)),$ where $\operatorname{sigmoid}(\cdot)$ is the activation function and $\hat{\mathcal{R}} \in \mathbb{R}^{hw \times hw}$. Then we refine $CAM_s$ from SVC into static pseudo labels $M_s$ and leverage its pixel-wise affinity to guide the diversifying of $\hat{\mathcal{R}}$. Specifically, if the pixel with coordinate $(i,j)$ shows the same pixel value as the pixel in $(\varepsilon, \eta)$, the token pair with the same coordinates on $F_d$ is semantically related and its corresponding correlation logit on $\hat{\mathcal{R}}$ should be maximized, and vice versa. The diversity loss can be formulated as:
\begin{equation}                           
\mathcal{L}_{\text {div }}=\frac{1}{{~N}^{+}} \sum_{{u}^{+} \in {\hat{\mathcal{R}}}^{+}}(1-{u}^{+})+\frac{1}{{~N}^{-}} \sum_{{u}^{-} \in {\hat{\mathcal{R}}}^{-}}{u}^{-},
    \label{eq:11}
\end{equation}
where $N^+/N^-$ are the number of positive and negative pairs, $u^{+}/u^{-}$ are the positive and negative relation logits, and $\hat{\mathcal{R}}^+/\hat{\mathcal{R}}^-$ are the positive and negative sets of logit on $\hat{\mathcal{R}}$, respectively. The diversity loss groups tokens with similar semantics and suppresses irrelevant ones, enhancing fine-grained details of visual features for precise text response. 

In addition, ExCEL is streamlined as a single-stage method. We adopt a lightweight Transformer-based segmentation head~\cite{18} and directly take the frozen visual encoder of CLIP for segmentation. The dynamic pseudo labels are used as supervision with a cross-entropy loss $\mathcal{L}_{seg}$. The loss objectives of our ExCEL are formulated as:
\begin{equation}                           
\mathcal{L}_{\text {ExCEL }}=\mathcal{L}_{{seg}}+\gamma \mathcal{L}_{\text {div}},
    \label{eq:12}
\end{equation}
where $\gamma$ is the weight factor. By efficiently training the adapter and a segmentation head, ExCEL achieves strong WSSS performance and significantly reduces training cost.

\section{Experiments and Results}
\subsection{Experimental Settings}

\textbf{Datasets and Metrics.} The proposed ExCEL is evaluated on PASCAL VOC 2012~\cite{VOC} and MS COCO 2014~\cite{COCO}. VOC contains $21$ categories ($1$ for background). Following prior methods~\cite{GSM,PPC,SeCo}, the augmented dataset with $10,582$, $1,449$, and $1,456$ images are used for training, validating, and testing, respectively. COCO includes $81$ classes, in which $82,081$ images are used for training and $40,137$ images are for validating. Mean Intersection-Over-Union (mIoU) is used as the main evaluation metric. 


\begin{table}[t]
\centering
\caption{Segmentation comparisons on VOC and COCO. Net. is the backbone for segmentation. Sup. is the supervision type. $\mathcal{I}$: image-level labels. $\mathcal{SA}$: saliency maps. $\mathcal{L}$: language. }
   \vspace{-1em}
    \tablestyle{3.7pt}{1}
    \scalebox{1.}
    {
    \footnotesize
    \begin{tabularx}{\linewidth}{@{}l|cccccc@{}}
    \toprule
    \multicolumn{1}{l|}{}                                                & \multicolumn{1}{c|}{}                                     & \multicolumn{1}{c|}{}                                       & \multicolumn{2}{c|}{VOC}                                                   & COCO          \\ \cline{4-6} 
    \multicolumn{1}{l|}{\multirow{-2}{*}{Method}}                        & \multicolumn{1}{c|}{\multirow{-2}{*}{Sup.}}                & \multicolumn{1}{c|}{\multirow{-2}{*}{Net.}}                 & Val           & \multicolumn{1}{c|}{Test}                                  & Val           \\ \midrule
    \multicolumn{6}{l}{\textit{\textbf{Multi-stage WSSS methods.}}}                                                                                                                                                                                                                             \\
    \multicolumn{1}{l|}{L2G~\cite{L2G} \tiny CVPR'2022}                                             & \multicolumn{1}{c|}{$\mathcal{I}+\mathcal{SA}$}                                  & \multicolumn{1}{c|}{RN101}                                  & 72.1          & \multicolumn{1}{c|}{71.7}                                  & 44.2          \\
    \multicolumn{1}{l|}{RCA~\cite{RCA} \tiny CVPR'2023}                                             & \multicolumn{1}{c|}{$\mathcal{I}+\mathcal{SA}$}                                  & \multicolumn{1}{c|}{RN38}                                   & 72.2          & \multicolumn{1}{c|}{72.8}                                  & 36.8          \\
    \multicolumn{1}{l|}{OCR~\cite{OCR} \tiny CVPR'2023}                                             & \multicolumn{1}{c|}{$\mathcal{I}$}                                    & \multicolumn{1}{c|}{RN38}                                   & 72.7          & \multicolumn{1}{c|}{72.0}                                  & 42.5          \\
    \multicolumn{1}{l|}{BECO~\cite{BECO} \tiny CVPR'2023}                                            & \multicolumn{1}{c|}{$\mathcal{I}$}                                    & \multicolumn{1}{c|}{RN101}                                  & 73.7          & \multicolumn{1}{c|}{73.5}                                  & 45.1          \\
    \multicolumn{1}{l|}{MCTformer+~\cite{33} \tiny TPAMI'2024}                                     & \multicolumn{1}{c|}{$\mathcal{I}$}                                    & \multicolumn{1}{c|}{RN38}                                   & 74.0          & \multicolumn{1}{c|}{73.6}                                  & 45.2          \\
    \multicolumn{1}{l|}{CTI~\cite{35} \tiny CVPR'2024}                                             & \multicolumn{1}{c|}{$\mathcal{I}$}                                    & \multicolumn{1}{c|}{RN101}                                  & 74.1          & \multicolumn{1}{c|}{73.2}                                  & 45.4          \\
    \multicolumn{1}{l|}{CLIMS~\cite{15} \tiny CVPR'2022}                                           & \multicolumn{1}{c|}{$\mathcal{I}+\mathcal{L}$}                                  & \multicolumn{1}{c|}{RN101}                                  & 70.4          & \multicolumn{1}{c|}{70.0}                                  & -             \\
    \multicolumn{1}{l|}{CLIP-ES~\cite{16} \tiny CVPR'2023}                                         & \multicolumn{1}{c|}{$\mathcal{I}+\mathcal{L}$}                                  & \multicolumn{1}{c|}{RN101}                                  & 72.2          & \multicolumn{1}{c|}{72.8}                                  & 45.4          \\
    \multicolumn{1}{l|}{PSDPM~\cite{PSDPM} \tiny CVPR'2024}                                           & \multicolumn{1}{c|}{$\mathcal{I}+\mathcal{L}$}                                  & \multicolumn{1}{c|}{RN101}                                  & 74.1          & \multicolumn{1}{c|}{74.9}                                  & 47.2          \\
    \multicolumn{1}{l|}{CPAL~\cite{37} \tiny CVPR'2024}                                            & \multicolumn{1}{c|}{$\mathcal{I}+\mathcal{L}$}                                  & \multicolumn{1}{c|}{RN101}                                  & 74.5          & \multicolumn{1}{c|}{74.7}                                  & 46.8          \\ \midrule
    \multicolumn{6}{l}{\textit{\textbf{Single-stage WSSS methods.}}}                                                                                                                                                                                                                            \\
    \multicolumn{1}{l|}{AFA~\cite{12} \tiny CVPR'2022}                                             & \multicolumn{1}{c|}{$\mathcal{I}$}                                    & \multicolumn{1}{c|}{MiT-B1}                                 & 66.0          & \multicolumn{1}{c|}{66.3}                                  & 38.9          \\
    \multicolumn{1}{l|}{ViT-PCM~\cite{ViT-PCM} \tiny ECCV'2022}                                         & \multicolumn{1}{c|}{$\mathcal{I}$}                                    & \multicolumn{1}{c|}{ViT-B}                                  & 70.3          & \multicolumn{1}{c|}{70.9}                                  & -             \\
    \multicolumn{1}{l|}{ToCo~\cite{22} \tiny CVPR'2023}                                            & \multicolumn{1}{c|}{$\mathcal{I}$}                                    & \multicolumn{1}{c|}{ViT-B}                                  & 71.1          & \multicolumn{1}{c|}{72.2}                                  & 42.3          \\
    \multicolumn{1}{l|}{DuPL~\cite{DuPL} \tiny CVPR'2024}                                            & \multicolumn{1}{c|}{$\mathcal{I}$}                                    & \multicolumn{1}{c|}{ViT-B}                                  & 73.3          & \multicolumn{1}{c|}{72.8}                                  & 44.6          \\
    \multicolumn{1}{l|}{SeCo~\cite{SeCo} \tiny CVPR'2024}                                            & \multicolumn{1}{c|}{$\mathcal{I}$}                                    & \multicolumn{1}{c|}{ViT-B}                                  & 74.0          & \multicolumn{1}{c|}{73.8}                                  & 46.7          \\
    \multicolumn{1}{l|}{DIAL~\cite{DIAL} \tiny ECCV'2024}                                            & \multicolumn{1}{c|}{$\mathcal{I}+\mathcal{L}$}                                  & \multicolumn{1}{c|}{ViT-B}                                  & 74.5          & \multicolumn{1}{c|}{74.9}                                  & 44.4          \\
    \multicolumn{1}{l|}{WeCLIP~\cite{18} \tiny CVPR'2024}                                          & \multicolumn{1}{c|}{$\mathcal{I}+\mathcal{L}$}                                  & \multicolumn{1}{c|}{ViT-B}                                  & 76.4          & \multicolumn{1}{c|}{77.2}                                  & 47.1          \\
    \rowcolor[HTML]{EFEFEF} 
    \multicolumn{1}{l|}{\cellcolor[HTML]{EFEFEF}\textbf{ExCEL(w/o CRF)}} & \multicolumn{1}{c|}{\cellcolor[HTML]{EFEFEF}\textbf{$\mathcal{I}+\mathcal{L}$}} & \multicolumn{1}{c|}{\cellcolor[HTML]{EFEFEF}\textbf{ViT-B}} & \textbf{77.2} & \multicolumn{1}{c|}{\cellcolor[HTML]{EFEFEF}\textbf{77.3}} & \textbf{49.3} \\
    \rowcolor[HTML]{EFEFEF} 
    \multicolumn{1}{l|}{\cellcolor[HTML]{EFEFEF}\textbf{ExCEL (Ours)}}          & \multicolumn{1}{c|}{\cellcolor[HTML]{EFEFEF}\textbf{$\mathcal{I}+\mathcal{L}$}} & \multicolumn{1}{c|}{\cellcolor[HTML]{EFEFEF}\textbf{ViT-B}} & \textbf{78.4} & \multicolumn{1}{c|}{\cellcolor[HTML]{EFEFEF}\textbf{78.5}} & \textbf{50.3} \\ \bottomrule
    \end{tabularx}
    }
   \label{tab.1}
   \vspace{-1em}
\end{table}


\begin{table}[t!]
\centering
\caption{CAM seed comparisons on VOC train set. $\mathcal{M}$: multi-stage methods. $\mathcal{S}$: single-stage methods. $\dagger$: our reproduction following official codes. ExCEL*: ExCEL in a training-free manner. }
   \vspace{-1em}
    \tablestyle{6.5pt}{1}
    \scalebox{1.}
    {
    \footnotesize
    \begin{tabular}{@{}lcccc}
    \toprule
    \multicolumn{1}{l|}{}                                        & \multicolumn{1}{c|}{}                                   & \multicolumn{1}{c|}{}                                     & \multicolumn{1}{c|}{}                                       & VOC           \\ \cline{5-5} 
    \multicolumn{1}{l|}{\multirow{-2}{*}{Method}}                & \multicolumn{1}{c|}{\multirow{-2}{*}{Type}}              & \multicolumn{1}{c|}{\multirow{-2}{*}{Sup.}}               & \multicolumn{1}{c|}{\multirow{-2}{*}{Net.}}                 & Train         \\ \midrule
    \multicolumn{5}{l}{\textit{\textbf{Training-free WSSS methods.}}}                                                                                                                                                                                                              \\
    \multicolumn{1}{l|}{CLIP-ES~\cite{16} \tiny CVPR'2023}                                 & \multicolumn{1}{c|}{$\mathcal{M}$}                                  & \multicolumn{1}{c|}{$\mathcal{I}+\mathcal{L}$}                                  & \multicolumn{1}{c|}{ViT-B}                                  & 70.8          \\
    \rowcolor[HTML]{EFEFEF} 
    \multicolumn{1}{l|}{\cellcolor[HTML]{EFEFEF}\textbf{ExCEL* (Ours)}} & \multicolumn{1}{c|}{\cellcolor[HTML]{EFEFEF}\textbf{$\mathcal{S}$}} & \multicolumn{1}{c|}{\cellcolor[HTML]{EFEFEF}\textbf{$\mathcal{I}+\mathcal{L}$}} & \multicolumn{1}{c|}{\cellcolor[HTML]{EFEFEF}\textbf{ViT-B}} & \textbf{74.6} \\ \midrule
    \multicolumn{5}{l}{\textit{\textbf{Training-required WSSS methods.}}}                                                                                                                                                                                                          \\
    \multicolumn{1}{l|}{ReCAM~\cite{30} \tiny CVPR'2022}                                   & \multicolumn{1}{c|}{$\mathcal{M}$}                                  & \multicolumn{1}{c|}{$\mathcal{I}$}                                    & \multicolumn{1}{c|}{RN101}                                  & 54.8          \\
    \multicolumn{1}{l|}{FPR~\cite{31} \tiny CVPR'2023}                                     & \multicolumn{1}{c|}{$\mathcal{M}$}                                  & \multicolumn{1}{c|}{$\mathcal{I}$}                                    & \multicolumn{1}{c|}{RN101}                                  & 63.8          \\
    \multicolumn{1}{l|}{LPCAM~\cite{32} \tiny CVPR'2023}                                 & \multicolumn{1}{c|}{$\mathcal{M}$}                                  & \multicolumn{1}{c|}{$\mathcal{I}$}                                    & \multicolumn{1}{c|}{RN50}                                   & 65.3          \\
    \multicolumn{1}{l|}{MCTformer+~\cite{33} \tiny TPAMI'2024}                              & \multicolumn{1}{c|}{$\mathcal{M}$}                                  & \multicolumn{1}{c|}{$\mathcal{I}$}                                    & \multicolumn{1}{c|}{RN38}                                   & 68.8          \\
    \multicolumn{1}{l|}{SFC~\cite{34} \tiny AAAI'2024}                                     & \multicolumn{1}{c|}{$\mathcal{M}$}                                  & \multicolumn{1}{c|}{$\mathcal{I}$}                                    & \multicolumn{1}{c|}{RN101}                                  & 64.7          \\
    \multicolumn{1}{l|}{CTI~\cite{35} \tiny CVPR'2024}                                     & \multicolumn{1}{c|}{$\mathcal{M}$}                                  & \multicolumn{1}{c|}{$\mathcal{I}$}                                    & \multicolumn{1}{c|}{RN101}                                  & 69.5          \\
    \multicolumn{1}{l|}{AFA~\cite{12} \tiny CVPR'2022}                                     & \multicolumn{1}{c|}{$\mathcal{S}$}                                  & \multicolumn{1}{c|}{$\mathcal{I}$}                                    & \multicolumn{1}{c|}{MiT-B1}                                 & 65.0          \\
    \multicolumn{1}{l|}{ViT-PCM~\cite{ViT-PCM} \tiny ECCV'2022}                                 & \multicolumn{1}{c|}{$\mathcal{S}$}                                  & \multicolumn{1}{c|}{$\mathcal{I}$}                                    & \multicolumn{1}{c|}{ViT-B}                                  & 67.7          \\
    \multicolumn{1}{l|}{$\dagger$ToCo~\cite{22} \tiny CVPR'2023}                                    & \multicolumn{1}{c|}{$\mathcal{S}$}                                  & \multicolumn{1}{c|}{$\mathcal{I}$}                                    & \multicolumn{1}{c|}{ViT-B}                                  & 71.6          \\
    \multicolumn{1}{l|}{$\dagger$DuPL~\cite{DuPL} \tiny CVPR'2024}                                    & \multicolumn{1}{c|}{$\mathcal{S}$}                                  & \multicolumn{1}{c|}{$\mathcal{I}$}                                    & \multicolumn{1}{c|}{ViT-B}                                  & 75.0          \\
    \multicolumn{1}{l|}{SeCo~\cite{SeCo} \tiny CVPR'2024}                                    & \multicolumn{1}{c|}{$\mathcal{S}$}                                  & \multicolumn{1}{c|}{$\mathcal{I}$}                                    & \multicolumn{1}{c|}{ViT-B}                                  & 74.8          \\
    \multicolumn{1}{l|}{CLIMS~\cite{15} \tiny CVPR'2022}                                   & \multicolumn{1}{c|}{$\mathcal{M}$}                                  & \multicolumn{1}{c|}{$\mathcal{I}+\mathcal{L}$}                                  & \multicolumn{1}{c|}{RN101}                                  & 56.6          \\
    \multicolumn{1}{l|}{POLE~\cite{36} \tiny WACV'2023}                                    & \multicolumn{1}{c|}{$\mathcal{M}$}                                  & \multicolumn{1}{c|}{$\mathcal{I}+\mathcal{L}$}                                  & \multicolumn{1}{c|}{RN50}                                   & 59.0          \\
    \multicolumn{1}{l|}{CPAL~\cite{37} \tiny CVPR'2024}                                    & \multicolumn{1}{c|}{$\mathcal{M}$}                                  & \multicolumn{1}{c|}{$\mathcal{I}+\mathcal{L}$}                                  & \multicolumn{1}{c|}{RN101}                                  & 71.9          \\
    \multicolumn{1}{l|}{DIAL~\cite{DIAL} \tiny ECCV'2024}                                    & \multicolumn{1}{c|}{$\mathcal{S}$}                                  & \multicolumn{1}{c|}{$\mathcal{I}+\mathcal{L}$}                                  & \multicolumn{1}{c|}{ViT-B}                                  & 75.2          \\
    \multicolumn{1}{l|}{$\dagger$WeCLIP~\cite{18} \tiny CVPR'2024}                                  & \multicolumn{1}{c|}{$\mathcal{S}$}                                  & \multicolumn{1}{c|}{$\mathcal{I}+\mathcal{L}$}                                  & \multicolumn{1}{c|}{ViT-B}                                  & 75.4          \\
    \rowcolor[HTML]{EFEFEF} 
    \multicolumn{1}{l|}{\cellcolor[HTML]{EFEFEF}\textbf{ExCEL (Ours)}}  & \multicolumn{1}{c|}{\cellcolor[HTML]{EFEFEF}\textbf{$\mathcal{S}$}} & \multicolumn{1}{c|}{\cellcolor[HTML]{EFEFEF}\textbf{$\mathcal{I}+\mathcal{L}$}} & \multicolumn{1}{c|}{\cellcolor[HTML]{EFEFEF}\textbf{ViT-B}} & \textbf{78.0} \\ \bottomrule
    \end{tabular}
    }
   \label{tab.2}
   \vspace{-1em}
\end{table}

\begin{figure*}[h!]
  \centering
  \includegraphics[width=17.4cm]{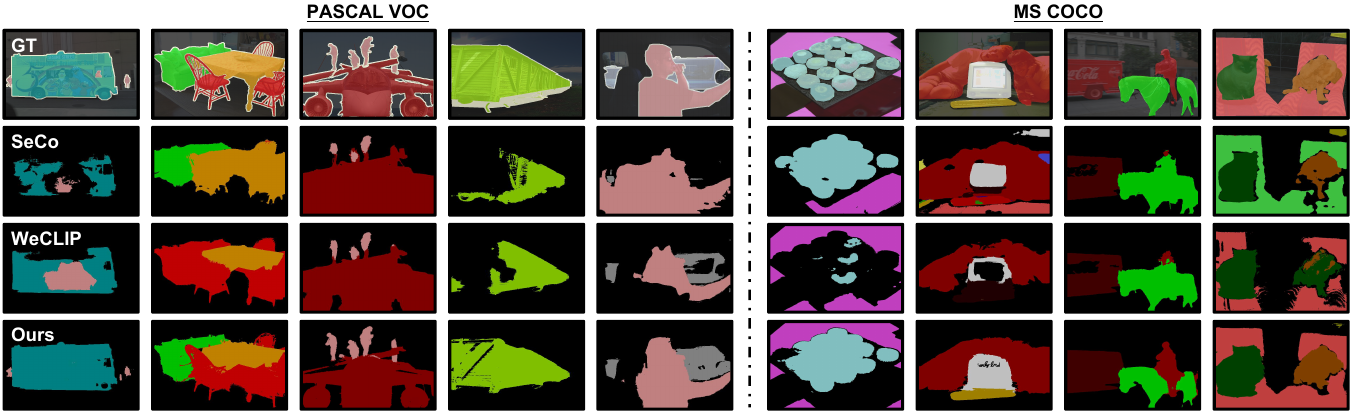}
  \vspace{-1.5em}
   \caption{Segmentation visualizations of SeCo~\cite{SeCo}, WeCLIP~\cite{18} and ours on VOC and COCO. ExCEL segments objects more precisely.}
   \label{fig.3}
   \vspace{-1.em}
\end{figure*}

\textbf{Implementation Details.} CLIP model with ViT-B~\cite{ViT} is used as ExCEL's encoder, which is frozen during the training. For TSE module, we generate $n=20$ descriptions from GPT-4 for each category. The number of attribute embeddings $B$ is set to $112$ and $224$ for PASCAL VOC and MS COCO, respectively. The SVC module is conducted in the last $N=5$ Transformer layers. Our decoder adopts a simple Transformer-based head~\cite{18}. Features $F_l$ from each layer of CLIP are sent to it for the segmentation predictions. The scaling and shifting factors in~\cref{eq:8} are set as $3.0$ and $1.0$, respectively. The loss weight $\gamma$ is set as $0.1$. Following~\cite{SeCo, DuPL, 22}, the AdamW optimizer is used for training the adapter and decoder. The learning rate is 1e-4 with a weight decay of 1e-2. The training iteration is set as $30,000$ for VOC and $100,000$ for COCO. Please refer to Supplementary Materials for more details.

\subsection{Comparisons with State-of-the-art Methods}

\begin{figure*}[tp!]
  \centering
  \includegraphics[width=17.4cm]{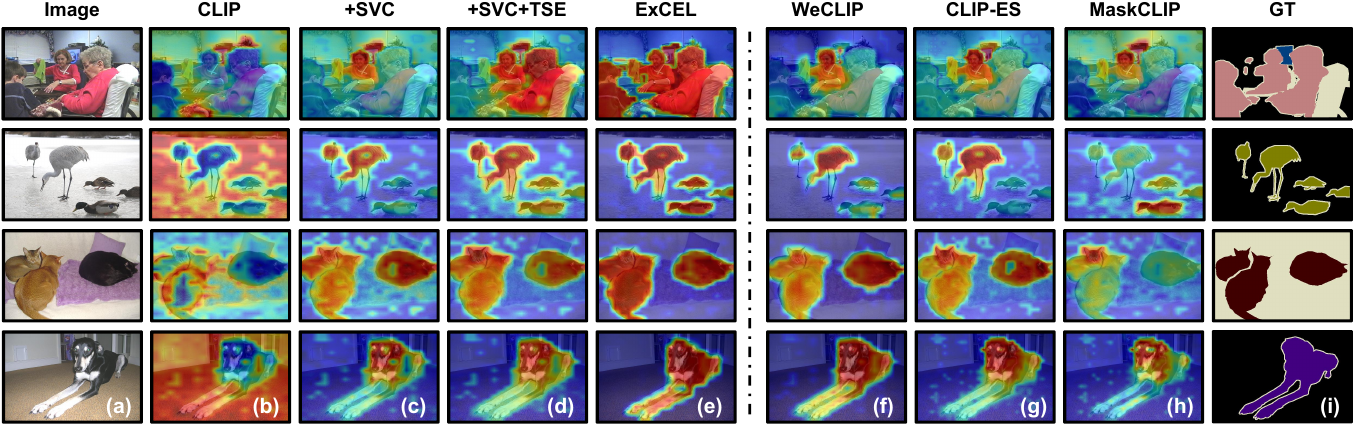}
  \vspace{-1em}
   \caption{CAM visualizations on VOC train set. (a) Image. (b-e) Ablative visualizations of proposed modules. (e-h) Qualitative comparisons of (e) ExCEL and recent CLIP-based methods, i.e., (f) WeCLIP~\cite{18}, (g) CLIP-ES~\cite{16} and (h) MaskCLIP~\cite{27}. (i) Ground truth.}
   \label{fig.4}
   \vspace{-1.em}
\end{figure*}

\textbf{Performance of Semantic Segmentation.} \cref{tab.1} shows segmentation comparisons between our ExCEL and recent methods on VOC and COCO. The single-stage ExCEL achieves $78.4\%$ and $78.5\%$ mIoU on VOC val set and test set, which even significantly outperforms the sophisticated multi-stage methods by at least $3.9\%$ and $3.6\%$ mIoU, respectively. For more complicated benchmark COCO, ExCEL achieves $50.3\%$ mIoU on val set, which brings a noticeable $3.2\%$ increase over the CLIP-based state-of-the-art (SOTA) WeCLIP. In addition, without time-consuming post-processing techniques, such as CRF~\cite{CRF}, ExCEL still maintains consistent superiority over SOTAs with CRF. 

The qualitative comparisons on VOC and COCO are shown in \cref{fig.3}. By densely matching the patches and texts, ExCEL consistently demonstrates more precise object segmentation than recent methods in an image-text paradigm.

\textbf{Evaluation of CAM Seeds.} \cref{tab.2} reports the quality of raw CAM seeds on VOC train set. Compared with recent methods, ExCEL achieves $74.6\%$ mIoU in a training-free setup, outperforming CLIP-ES by $3.8\%$ and performing comparably to most training-required methods. With the optimized LVC module, ExCEL further boosts CAM quality to $78.0\%$, surpassing SOTAs in the image-text paradigm by at least $2.6\%$. In addition, visual comparisons in \cref{fig.4} (e-h) also plainly illustrates that ExCEL generates better CAMs with the designed patch-text alignment paradigm.

\begin{table}[t!]
\centering
\caption{Ablation study of ExCEL on VOC val set.}
   \vspace{-1em}
    \tablestyle{5.4pt}{1}
    \scalebox{1.}
    {
    \footnotesize
    \begin{tabularx}{\linewidth}{@{}l|cccccc@{}}
    \toprule
    Conditions      & SVC         & TSE         & LVC         & Precision     & Recall        & mIoU          \\ \midrule
    Baseline (CLIP) &             &             &             & 18.8          & 21.3          & 12.1          \\
    w/ SVC           & \pmb{$\checkmark$}           &             &             & 81.2          & 86.2          & 72.5          \\
    w/o LVC          & \pmb{$\checkmark$}           & \pmb{$\checkmark$}             &             & 80.7          & 89.8          & 74.7          \\
    w/o TSE           & \pmb{$\checkmark$}           &             & \pmb{$\checkmark$}            & 83.7          & 86.3          & 75.1          \\
    \rowcolor[HTML]{EFEFEF} 
    \textbf{ExCEL}  & \textbf{\pmb{$\checkmark$}} & \textbf{\pmb{$\checkmark$}} & \textbf{\pmb{$\checkmark$}} & \textbf{85.0} & \textbf{88.4} & \textbf{77.2} \\ \bottomrule
    \end{tabularx}
    }
   \label{tab.3}
   \vspace{-0.5em}
\end{table}


\begin{table}[t!]
\centering
\caption{Ablation study of attribute number $B$ on VOC val set.}
   \vspace{-1em}
    \tablestyle{8.pt}{1}
    \scalebox{1.}
    {
    \footnotesize
    \begin{tabularx}{\linewidth}{@{}l|cccccc@{}}
    \toprule
    Number of Attr & None & 32 & 64 & \textbf{112}  & 144 & 196 \\ \midrule
    mIoU           & 75.1   &75.8    &76.2    & \textbf{77.2} & 77.0    & 76.5    \\ \bottomrule
    \end{tabularx}
    }
   \label{tab.4}
   \vspace{-0.5em}
\end{table}


\begin{table}[t!]
\centering
\caption{Ablation study of VC module on VOC train set.}
   \vspace{-1em}
    \tablestyle{3.4pt}{1}
    \scalebox{1.}
    {
    \footnotesize
    \begin{tabularx}{\linewidth}{@{}l|cccccccc@{}}
    \toprule
    Conditions & q-k & v & I.C. & \multicolumn{1}{l}{M.C.} & LVC & \multicolumn{1}{c}{Precision} & \multicolumn{1}{c}{Recall} & mIoU \\ \midrule
    Baseline (CLIP)   & \pmb{$\checkmark$}   &   &      &                          &     &18.0                               &21.8                            & 11.2 \\
    MaskCLIP   &     & \pmb{$\checkmark$} &      &                          &     &77.1                              &80.9                            & 65.8 \\
    w/ I.C.      &     &   & \pmb{$\checkmark$}    &                          &     &79.1                               &84.7                           & 69.7 \\
    SVC      &     &   & \pmb{$\checkmark$}    & \pmb{$\checkmark$}                        &     &82.2                               &88.2                           & 74.6 \\
    \rowcolor[HTML]{EFEFEF} 
    \textbf{ExCEL}      &     &   & \textbf{\pmb{$\checkmark$}}    & \textbf{\pmb{$\checkmark$}}                        & \textbf{\pmb{$\checkmark$}}   &\textbf{86.6}                               &\textbf{87.9}                            & \textbf{78.0} \\  \bottomrule
    \end{tabularx}
    }
   \label{tab.5}
   \vspace{-1.5em}
\end{table}
\subsection{Ablation Studies}

\textbf{Efficacy of Key Components.} Quantitative ablative experiments of ExCEL are reported in \cref{tab.3}. The baseline is the vanilla CLIP using our training settings, which only achieves $12.1\%$ mIoU for segmentation. Our SVC module replaces the q-k attention with Intra-correlation from the intermediate layers. The performance increases to $72.5\%$. TSE module enriches the semantics of text representation for robust visual recognition. Introducing TSE brings a $3.6\%$ recall increase compared to the original text templates. LVC module provides a dynamic shift to diversify the features. It further benefits SVC's segmentation performance to $75.1\%$ mIoU. With all these enhancements, ExCEL generates $77.2\%$ mIoU for segmentation.

Qualitative ablation results are further illustrated in \cref{fig.4} (b-e) to evaluate the efficacy of our modules. In \cref{fig.4} (b), the CLIP baseline produces inaccurate CAMs with mislocalized activation. SVC corrects token relations and preserves fine-grained details, effectively suppressing false activations, as seen in \cref{fig.4} (c). TSE incorporates comprehensive textual attributes into the text representation, enhancing patch-text matching and producing more complete CAMs, shown in \cref{fig.4} (d). LVC dynamically optimizes attention maps, further improving CAM accuracy and completeness, as illustrated in \cref{fig.4} (e). Both quantitative and qualitative results confirm the effectiveness of our modules.

\textbf{Effectiveness of Implicit Attributes.} \cref{tab.4} analyzes the effect of varying the number of clustering attributes. 'None' means no clustering and we explicitly fuse the $n$ description embeddings for each class. It shows that the performance drops from $77.2\%$ to $75.1\%$, which validates the efficacy of implicit attributes and its superiority over explicit descriptions. With this operation, we can expand the representation of $20$ classes up to $196$ attributes or more, greatly enhancing text semantics. It reports that ExCEL achieves more favorable performance when $B$ is set to $112$.

\textbf{Effectiveness of Visual Calibrations.} \cref{tab.5} compares different strategies in VC module for CAM generation. I.C. refers to Intra-correlation in the last layer, and M.C. (Intermediate Calibration) applies I.C. across intermediate layers. Vanilla q-k attention in CLIP loses diversity and cannot generate reasonable CAMs. $v$ contains fine-grained knowledge and MaskCLIP improves CAMs to $65.8\%$ by using $v$ from the last layer. In contrast, our I.C. and M.C. focus on mining diverse knowledge from intermediate layers and boost the performance to $69.7\%$ and $74.6\%$. In addition, introducing LVC module raises the final performance to $78.0\%$. Results in \cref{tab.5} and corresponding visualizations in \cref{fig.4} clearly highlight the efficacy of our components. 

\subsection{Further Analysis}
\label{sec.4.4}

\begin{table}[!t]
    \centering
    \caption{Comparisons with the fully-supervised counterparts on VOC val set. $\mathcal{F}$:fully-supervised. ViT-B*: pretrained from CLIP.}
    \vspace{-1em}
    \tablestyle{4.8pt}{1.1}
    \scalebox{1.}
    {
    \footnotesize
    \begin{tabularx}{\linewidth}{@{}l|c|c|c|cc@{}}
    \toprule
    Methods        & Type                   & Sup.                                                                            & Net.                                                            & Val           & Ratio           \\ \midrule
    {DeepLabV2~\cite{DeepLabV2} \tiny TPAMI'2017}      & -                      & \multicolumn{1}{c|}{$\mathcal{F}$}                                              & \multicolumn{1}{c|}{RN101}                                      & 77.7          & -               \\
    {DeepLabV2~\cite{DeepLabV2} \tiny TPAMI'2017}      & -                      & \multicolumn{1}{c|}{$\mathcal{F}$}                                              & \multicolumn{1}{c|}{ViT-B}                                   & 82.3          & -               \\
    {WeCLIP-Full~\cite{18} \tiny CVPR'2024}     & -                      & \multicolumn{1}{c|}{$\mathcal{F}$}                                              & \multicolumn{1}{c|}{ViT-B*}                                  & 81.6          & -               \\ \midrule
    {CLIMS~\cite{15} \tiny CVPR'2022}          & $\mathcal{M}$          & \multicolumn{1}{c|}{$\mathcal{I}+\mathcal{L}$}                                  & \multicolumn{1}{c|}{RN101}                                      & 70.4          & 90.6\%          \\
    {CLIP-ES~\cite{16} \tiny CVPR'2023}       & $\mathcal{M}$          & \multicolumn{1}{c|}{$\mathcal{I}+\mathcal{L}$}                                  & \multicolumn{1}{c|}{RN101}                                      & 72.2          & 92.9\%          \\
    {CPAL~\cite{37} \tiny CVPR'2024}          & $\mathcal{M}$          & \multicolumn{1}{c|}{$\mathcal{I}+\mathcal{L}$}                                  & \multicolumn{1}{c|}{RN101}                                      & 74.5          & 95.9\%          \\
    {ToCo~\cite{22} \tiny CVPR'2024}         &$\mathcal{S}$       & \multicolumn{1}{c|}{$\mathcal{I}$}     & \multicolumn{1}{c|}{ViT-B}           & 71.1          & 86.4\%          \\
    {DuPL~\cite{DuPL} \tiny CVPR'2024}        & $\mathcal{S}$      & \multicolumn{1}{c|}{$\mathcal{I}$}    & \multicolumn{1}{c|}{ViT-B}                                   & 73.3          & 89.1\%          \\
    {SeCo~\cite{18} \tiny CVPR'2024}           & $\mathcal{S}$          & \multicolumn{1}{c|}{$\mathcal{I}$}                                              & \multicolumn{1}{c|}{ViT-B}                                   & 74.0          & 89.9\%          \\
    {DIAL~\cite{DIAL} \tiny ECCV'2024}            & $\mathcal{S}$          & \multicolumn{1}{c|}{$\mathcal{I}+\mathcal{L}$}                                  & \multicolumn{1}{c|}{ViT-B}                                   & 74.5          & 90.5\%          \\
    {WeCLIP~\cite{18} \tiny CVPR'2024}         & $\mathcal{S}$          & \multicolumn{1}{c|}{$\mathcal{I}+\mathcal{L}$}                                  & \multicolumn{1}{c|}{ViT-B*}                                  & 76.4          & 93.6\%          \\
    \rowcolor[HTML]{EFEFEF} 
    {\cellcolor[HTML]{EFEFEF}\textbf{ExCEL (Ours)}} & \textbf{$\mathcal{S}$} & \multicolumn{1}{c|}{\cellcolor[HTML]{EFEFEF}\textbf{$\mathcal{I}+\mathcal{L}$}} & \multicolumn{1}{c|}{\cellcolor[HTML]{EFEFEF}\textbf{ViT-B*}} & \textbf{78.4} & \textbf{96.1\%} \\  \bottomrule
    \end{tabularx}
    }
   \label{tab.6}
   \vspace{-1em}
\end{table}

\begin{table}[!t]
    \centering
    \caption{Training efficiency comparisons on VOC train (CAM) and val set (Seg). All experiments are conducted on RTX 3090.}
    \vspace{-1em}
    \tablestyle{2.9pt}{1.14}
    \scalebox{1.}
    {
    \footnotesize
    \begin{tabularx}{\linewidth}{@{}l|ccccc@{}}
    \toprule
    Method                                  & Type                   & Training Time                               & GPU                                    & CAM                                   & Seg                                \\ \midrule
    {CLIMS~\cite{15} \tiny CVPR'2022}                                   & $\mathcal{M}$          & 1068 mins                          & 18.0 G                                   & 56.6                                  & 70.4                               \\
    {CLIP-ES~\cite{16} \tiny CVPR'2023}                                 & $\mathcal{M}$          & 420 mins                           & 12.0 G                                   & 70.8                                  & 72.2                               \\
    {MCTformer+~\cite{33} \tiny TPAMI'2024}                              & $\mathcal{M}$          & 1496 mins                          & 18.0 G                                   & 68.8                                  & 74.0                               \\
    {ToCo~\cite{22} \tiny CVPR'2023}                                    & $\mathcal{S}$          & 506 mins                           & 17.9 G                                 & 71.6                                  & 71.1                               \\
    {DuPL~\cite{DuPL} \tiny CVPR'2024}                                    & $\mathcal{S}$          & 508 mins                           & 14.9 G                                 & 75.0                                  & 73.3                               \\
    {SeCo~\cite{SeCo} \tiny CVPR'2024}                                    & $\mathcal{S}$          & 407 mins                           & 17.6 G                                 & 74.8                                  & 74.0                               \\
    {WeCLIP~\cite{18} \tiny CVPR'2024}                                  & $\mathcal{S}$          & 270 mins                           & 6.2 G                                  & 75.4                                  & 76.4                               \\
    \rowcolor[HTML]{EFEFEF} 
    {\cellcolor[HTML]{EFEFEF}\textbf{ExCEL* (Training-free)}} & \textbf{$\mathcal{S}$}                      & \textbf{-}       & \textbf{2.9 G} & \textbf{74.6} & \textbf{-}    \\
    \rowcolor[HTML]{EFEFEF} 
    {\cellcolor[HTML]{EFEFEF}\textbf{ExCEL (Ours)}}                          & \textbf{$\mathcal{S}$} & \textbf{90 mins}                   & \textbf{3.2 G}                         & \textbf{78.0}    & \textbf{78.4}                      \\ \bottomrule
    \end{tabularx}
    }
   \label{tab.7}
   \vspace{-1.5em}
\end{table}

\textbf{Hyper-parameter Analysis.} Hyper-parameters, such as \textit{TOP-K, scaling factors $\alpha$ and $\beta$, and the number of SVC layers $N$, etc.}, are discussed in \textbf{Supplementary Materials}.

\textbf{Fully-supervised Counterparts.} \cref{tab.6} presents a fairness comparison between WSSS methods and their fully-supervised counterparts using the same segmentation backbone. With CLIP's visual encoder as the backbone, ExCEL achieves $78.4\%$ mIoU, reaching $96.1\%$ of the fully-supervised performance. It significantly outperforms CLIP-based WeCLIP by $2.5\%$ and demonstrates ExCEL's advantage over other multi-stage CLIP-based methods as well.

\textbf{Training Efficiency Analysis.} 
Our method only trains the adapter and decoder in a single-stage paradigm. \cref{tab.7} compares the training efficiency between ExCEL and recent methods. Without training, ExCEL requires just $2.9$ GB of GPU memory and generates comparable CAMs to recent SOTAs. When training is included, the entire pipeline only takes $90$ minutes and $3.2$ GB of memory for SOTA performance. ExCEL just requires \textbf{\textcolor{teal}{$6.0\%$}} training time of multi-stage MCTformer+ and \textbf{\textcolor{teal}{$33.3\%$}} of single-stage WeCLIP, highlighting ExCEL's remarkable training efficiency.  

\begin{figure}[!t]
  \centering
  \includegraphics[width=8.2cm]{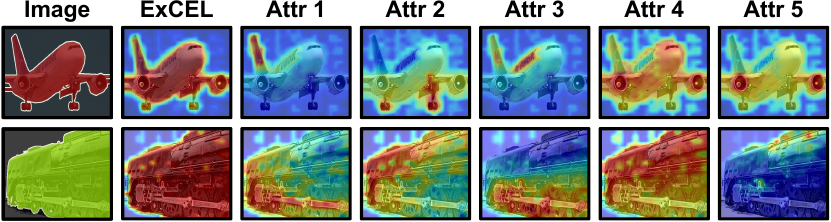}
   \vspace{-.5em}
   \caption{Implicit attribute responses. Based on the TOPK similarity scores, 5 attributes are sampled for visualizations. }
   \label{fig.5}
   \vspace{-0.8em}
\end{figure}

\textbf{Attribute Response Analysis.} We treat text prompting as an implicit attribute-hunting process to comprehensively enrich text representation semantics. To evaluate if the clustered attributes capture distinct object characteristics, we visualize $5$ implicit attributes based on similarity scores. As shown in \cref{fig.5}, given instances of $\{aeroplane\}$ and $\{train\}$, our attributes highlight different object parts, which clearly validates that our TSE module enhances integral visual responses by gathering relevant semantics.

\begin{figure}[!t]
  \centering
  \includegraphics[width=8.4cm]{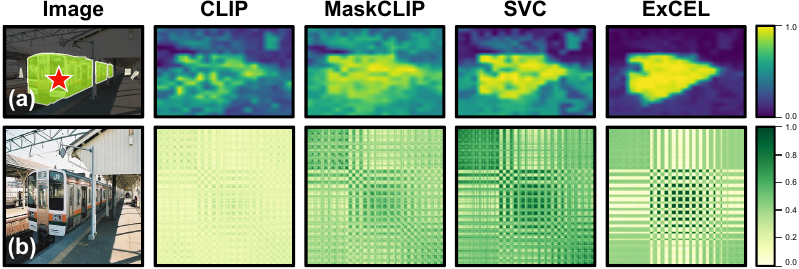}
   \vspace{-1.6em}
   \caption{Comparisons of attention maps from the last visual encoder layer. (a) Attention features from the query patches (marked by red stars). (b) Token relations measured by cosine similarity.}
   \label{fig.6}
   \vspace{-1.5em}
\end{figure}

\textbf{Feature Representation Analysis.} CLIP lacks fine-grained details, leading to inaccurate patch-text responses. To explore further, we visualize the self-attention features in \cref{fig.6} (a). Given the query patch (red star), CLIP's q-k attention falls short in generating diverse features, supporting our claim in \cref{sec.3.4}. MaskCLIP observes that $v$ keeps diversity and takes it from the last layer for visual response. We visualize it by calculating v-v attention. Although effective, MaskCLIP still misses fine granularity. Instead, SVC calculates attention within each space and implements it from intermediate layers. LVC further diversifies the features with a dynamic adapter, both of which effectively generate features with clear boundaries and spatial details. 

Additionally, we explore the pairwise token relations in \cref{fig.6} (b). Unlike the smoother attention maps of CLIP or MaskCLIP, our approach distinctly groups tokens with similar semantics, aligning pairwise similarities with corresponding semantics. This validates that ExCEL successfully enhances the frozen features of CLIP by calibrating it towards distributions with more diverse spatial information.

\section{Conclusion}
In this paper, we propose ExCEL, a novel patch-text alignment method to explore CLIP's dense knowledge for WSSS, which provides a different insight to generate better pseudo labels based on CLIP. To this end, Text Semantic Enrichment (TSE) and Visual Calibration (VC) modules are designed to improve the dense alignment across text and vision modalities. In addition, ExCEL generates CAMs in both training-free and efficient training modes, calibrating CLIP without altering its pre-trained weights. It retains CLIP's transferability while significantly reducing training cost. We believe ExCEL can inspire more future research to unlock CLIP's dense capabilities in the WSSS field.

\section{Acknowledgments}
This work was supported by the National Natural Science Foundation of China under Grant No.82372097, Shanghai Sailing Program under Grant 22YF1409300, International Science and Technology Cooperation Program under the 2023 Shanghai Action Plan for Science under Grant 23410710400, Taishan Scholars Program under Grant NO.tsqn202408245.

{
    \small
    \bibliographystyle{ieeenat_fullname}
    \bibliography{main}
}


\end{document}